\documentclass[11pt]{article}

\usepackage[preprint]{acl}

\usepackage{times}
\usepackage{latexsym}

\usepackage[T1]{fontenc}

\usepackage[utf8]{inputenc}

\usepackage{microtype}

\usepackage{inconsolata}

\usepackage{graphicx}
\usepackage{amsmath}
\usepackage{amsthm}
\usepackage{booktabs}
\usepackage{algorithm}
\usepackage{algorithmic}
\usepackage{tabularx}   
\usepackage{comment}
\usepackage{multirow}
\usepackage[switch]{lineno}
\usepackage{enumitem}
\title{WarrantScore: Modeling Warrants between Claims and Evidence for Substantiation Evaluation in Peer Reviews}

\author{
  \textbf{Kiyotada Mori\textsuperscript{1,4,*}},
  \textbf{Shohei Tanaka\textsuperscript{2}},
  \textbf{Tosho Hirasawa\textsuperscript{2}},
  \textbf{Tadashi Kozuno\textsuperscript{2}},
\\
  \textbf{Koichiro Yoshino\textsuperscript{3,4,1}},
 \textbf{Yoshitaka Ushiku\textsuperscript{2}}
\\
\\
  \textsuperscript{1}Nara Institute of Science and Technology,
  \textsuperscript{2}OMRON SINIC X Corporation,
  \\
  \textsuperscript{3}Institute of Science Tokyo,
  \textsuperscript{4}Guardian Robot Project, RIKEN,
}

\begin{document}
\maketitle

\renewcommand{\thefootnote}{\fnsymbol{footnote}} 
\footnotetext[1]{Work done during an internship at OMRON SINIC X Corporation.}

\begin{abstract}
The scientific peer-review process is facing a shortage of human resources due to the rapid growth in the number of submitted papers.
The use of language models to reduce the human cost of peer review has been actively explored as a potential solution to this challenge. A method has been proposed to evaluate the level of substantiation in scientific reviews in a manner that is interpretable by humans. This method extracts the core components of an argument, claims and evidence, and assesses the level of substantiation based on the proportion of claims supported by evidence. The level of substantiation refers to the extent to which claims are based on objective facts. However, when assessing the level of substantiation, simply detecting the presence or absence of supporting evidence for a claim is insufficient; it is also necessary to accurately assess the logical inference between a claim and its evidence. We propose a new evaluation metric for scientific review comments that assesses the logical inference between claims and evidence.  Experimental results show that the proposed method achieves a higher correlation with human scores than conventional methods, indicating its potential to better support the efficiency of the peer-review process.
\end{abstract}

\section{Introduction}
The rapid development of scientific research has raised concerns about the increasing human costs associated with the peer-review process. Leading international conferences in computer science have seen a sharp increase in the number of papers submitted year-by-year. However, the number of reviewers has not matched the increase in the number of papers submitted \cite{tran2020open}. Due to the shortage of reviewers, the burden on conference organizers of ensuring the quality of reviews continues to increase \cite{stelmakh2021novice}. 

A promising approach to address these issues is language model-assisted review. For example, automatic detection of low-quality reviews by language models may encourage reviewers to improve their reviews or support meta-reviewers in making decisions. For these purposes, it is important not only to assign a quality score to a review, but also to provide an evaluation process in a form that is easily understandable to humans.

As an evaluation metric that is easily understood by humans, SubstanScore \cite{guo2023automatic} has been proposed. SubstanScore evaluates the level of substantiation by extracting claims and evidence from reviews and measuring the proportion of claims supported by evidence. SubstanScore not only provides the final evaluation score but also generates descriptions identifying the claims and supporting evidence within the review. Descriptions of claims and evidence allow for the assessment of the level of substantiation in scientific reviews in a way that is interpretable by humans.

SubstanScore evaluates the level of substantiation by focusing only on claims and evidence, which are two of the three main components of Toulmin's model \cite{toulmin2003uses}, a well-established argument structure. However, the presence of claim and evidence pairs alone does not ensure a comprehensive evaluation of the quality of review comments. For example, consider the claim ``I will reject the submitted paper'' paired with the evidence ``The main topic of the paper has already been thoroughly examined in many studies.'' The logical inference between claims and evidence within a review is not always sufficient because the remaining core element of Toulmin's model, the warrant such as ``topics on which many studies have been conducted are not accepted'' which serves to logically connect the claim and the evidence, may be somewhat credible but is not always entirely convincing. Therefore, evaluating the level of substantiation requires not only checking for the presence of supporting evidence but also accurately assessing the causal connection between the claim and its supporting evidence.

\begin{figure}[t]
    \centering
    \includegraphics[width=1\linewidth]{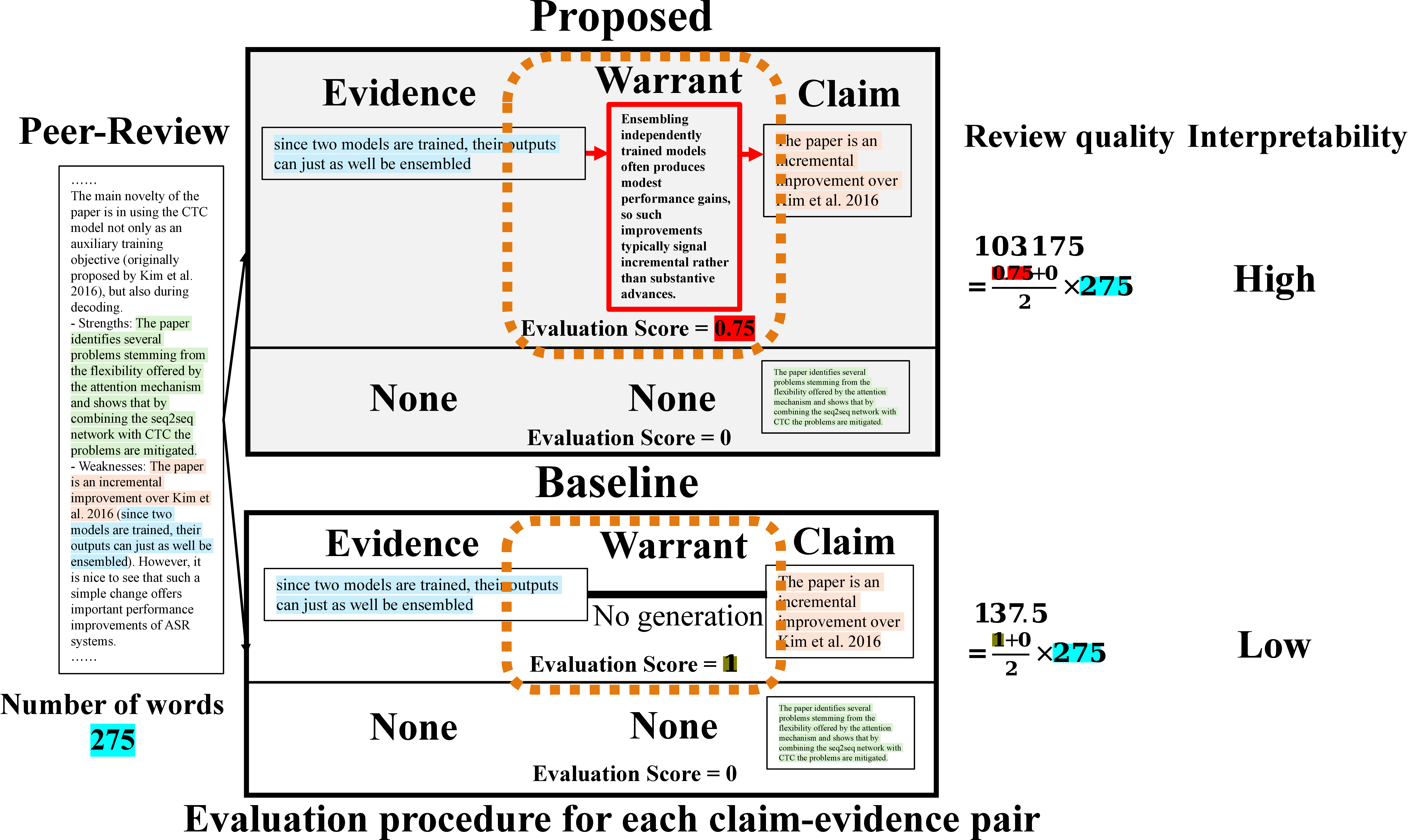} 
    \caption{Comparison between proposed peer-review evaluation metrics that evaluate claims, evidence, and warrants and baseline metrics that only evaluate claims and evidence.}
    \label{fig:concept}
\end{figure}

Inspired by Toulmin's model, we proposed a new evaluation metric for scientific review comments called WarrantScore as an extension of SubstanScore alone, by explicitly generating and evaluating warrants. Figure~\ref{fig:concept} illustrates this concept. 

We compared WarrantScore and SubstanScore using datasets annotated with subjective evaluations of humans  of the level of substantiation, namely SubstanReview \cite{guo2023automatic} and RottenReview \cite{ebrahimi2025rottenreviews}. The comparison was based on the correlation with subjective evaluations of humans. The results demonstrated that WarrantScore exhibits a stronger correlation with subjective evaluations of humans  than SubstanScore. Moreover, we confirmed the superiority of WarrantScore over regression models using subjective evaluations of humans as the target variable. To support the superiority of WarrantScore, we analyzed the ability of the evaluation metrics to assess the quality of scientific reviews, regardless of superficial linguistic features such as the number of words in a review. Each experiment is described in Section~\ref{sec:corr} and Section~\ref{sec:robust}. 

Our research addresses the challenge of automatically assessing the quality of scientific reviews in a way that is independent of superficial linguistic features and interpretable to humans.

\section{Related Work}
In this section, we first discuss the existing automatic evaluation metrics for scientific peer review comments in Section \ref{sec:eval_review} and describe the automatic extraction of argument components in Section \ref{sec:argumining}.

\subsection{Evaluation of Peer Review Comments}
\label{sec:eval_review}
Various evaluation metrics targeting specific aspects of scientific review comments have been proposed. For example, some focus on aspects such as harshness \cite{verma2022lack}, thoroughness and helpfulness \cite{severin2022journal}, or comprehensiveness \cite{yuan2022can}. However, these metrics are regression-based approaches that aim to predict subjective evaluations of human as the target variable. Therefore, even if these metrics provide an evaluation score for a given scientific review comment, humans cannot understand why that specific score was assigned.

SubstanScore attempts to automatically extract claims and evidence from scientific review comments and evaluate the level of substantiation based on the proportion of claims supported by evidence. Unlike regression methods, SubstanScore not only provides the final evaluation score but also shows, in a way that humans can verify, how the score was computed through the evaluation process. Whether the extracted evidence is truly correct and whether the evidence logically supports the claims remains unverified. Ryu et al. \cite{ryu2025reviewscore} proposed extracting the argumentative structure of scientific reviews using large language models (LLMs) to detect erroneous criticisms and verified the factual correctness of the evidence using LLMs. However, in the context of scientific review evaluation, no established method exists even now to verify that the evidence logically supports the claims.

\subsection{Argument Mining}
\label{sec:argumining}
An attempt has been made in the general domain to automatically extract argumentative components, such as warrants, from text using LLMs \cite{gupta2024harnessing}. In their study, GPT-4 \cite{achiam2023gpt} was prompted to generate argument components of Toulmin's model by adding the phrase ``According to Toulmin model'' to the input text. As a result, the warrants generated by GPT-4 were rated acceptable by two domain experts in 61.7\% of cases. They consider a warrant acceptable if it is: a) relevant and fully explains the link between the claim-reason pair, b) not trivial, c) must hold for the claim to be inferred from the reason, even if it does not align with the annotator's or reader's personal beliefs.

Since two experts rated 45.5\% of the warrants generated by humans as acceptable, it is suggested that sufficiently parameterized LLMs such as GPT-4 can generate warrants that humans may evaluate more favorably than human-authored warrants. Their study also has made available a dataset includes a total of 450 warrants along with binary acceptability labels assigned by two experts. The insights and resources released demonstrate the potential to generate and evaluate implicit warrants in scientific reviews.

\section{Problem Formulation}
In this section, we formalize the task of generating warrants, defined as text that establishes the logical inference between claims and evidence, which our proposed method is designed to address.

First, the set of claims extracted from a scientific review comment by humans or LLMs is defined as $C = \{ c_1, c_2, \ldots, c_N \}$, and similarly, the set of supporting evidence for these claims is defined as $E = \{ e_1, e_2, \ldots, e_N \}$. Each $c$ and $e$ represents text indicating a claim and its corresponding evidence within the scientific review comment. Elements of the set $C $ are all non $\emptyset$, but elements of $E$ may be $\emptyset$, reflecting that in scientific review comments, a claim does not always have supporting evidence.

Let $w$ be defined as text that logically connects $c$ and $e$; this text is called the warrant. The set of warrants is $W = \{ w_1, w_2, \ldots, w_N \}$. When $e$ is $\emptyset$, then $w$ is also $\emptyset$. Let LM denote the LLM as a function and let $p$ denote the instruction prompt. The relationship among $C$, $E$, and $W$ is formalized in Equation~(\ref{eq:warrant}).

\begin{equation}
\label{eq:warrant}
\begin{aligned}
W &= \{ w_1, w_2, \cdots, w_N \} \\
  &= \big\{
      \text{LM}(p, c_1, e_1),\; \text{LM(}p, c_2, e_2), \\
  &\qquad \ldots,\, \text{LM}(p, c_N, e_N)
     \big\} .
\end{aligned}
\end{equation}

Our proposed method assesses the level of substantiation in a scientific review comment by evaluating $W$ generated by LLM.

\section{Proposed Method}
In this section, we first describe the baseline method in Section \ref{sec:baseline}, then define the proposed method in Section \ref{sec:warrantscore} and finally detail the algorithm of our proposed method in Section \ref{sec:warrrant_gene_eval}.

\subsection{SubstanScore (Baseline)}
\label{sec:baseline}
Guo et al. \cite{guo2023automatic} proposed SubstanScore, an evaluation metric for the level of substantiation in scientific review comments. SubstanScore is calculated by multiplying the proportion of claims with supporting evidence by the number of words in the review. The definition of SubstanScore is shown in Equation (\ref{eq:substan-score}).

\begin{equation}
\label{eq:substan-score}
\begin{aligned}
& \text{SubstanScore} \\
& = \frac{\sum_{n=1}^{N} \text{support\_score}(c_n,e_n)}{|C|} \\
& \times \text{len}(\text{review}) \\
& = \text{supported\_claims} \times \text{len}(\text{review}). \\
\end{aligned}
\end{equation}

len(review) indicates the number of words in a scientific review comment, and $\text{support\_score}(c, e)$ is a function that returns 0 if $e$ is $\emptyset$, and 1 otherwise. supported\_claims represents the proportion of claims supported by evidence, calculated as
\begin{equation}
\label{eq:substan-rate}
\begin{aligned}
& \text{supported\_claims} \\
& = \frac{\sum_{n=1}^{N} \text{support\_score}(c_n, e_n)}{|C|}.
\end{aligned}
\end{equation}
SubstanScore does not verify whether the supporting evidence truly justifies the claim.

\subsection{WarrantScore}
\label{sec:warrantscore}
We proposed WarrantScore, an evaluation metric for scientific review comments that generates warrants, which are texts that logically connect claims and supporting evidence, and assess the plausibility of warrants, as an extension of SubstanScore. The definition of WarrantScore is expressed in Equation (\ref{eq:warrant-score}).

\begin{equation}
\label{eq:warrant-score}
\begin{aligned}
&\text{WarrantScore} \\
&= \dfrac{
      \sum_{n=1}^{N} \text{support\_score}(c_n, e_n,w_n)
   }{|C|} \\ &\times \text{len}(\text{review})\\
&= \dfrac{
      \sum_{n=1}^{N}
      \text{V}(w_n) \times \text{support\_score}(c_n, e_n)
   }{|C|} \\
& \times \text{len}(\text{review}) \\
& = \text{warrant\_rate} \times \text{len}(\text{review}). \\
& \quad \text{* note that } w_n = \text{LM}(p,c_n,e_n).
\end{aligned}
\end{equation}

WarrantScore evaluates both the presence of supporting evidence for claims and the plausibility of the warrant connecting each claim–evidence pair. In WarrantScore, $\text{support\_score}(c, e, w)$ is a function that returns 0 if $e = \emptyset$, and otherwise returns an evaluation score for $w$ in the range from 0 to 1. The warrant\_rate represents the mean evaluation score of warrants, calculated as

\begin{equation}
\label{eq:warrant-rate}
\begin{aligned}
& \text{warrant\_rate} \\
& = \frac{
\sum_{n=1}^{N} \text{V}(w_n) \times \text{support\_score}(c_n, e_n)
}{|C|}.
\end{aligned}
\end{equation}

SubstanScore can be interpreted as a special case of WarrantScore, in which $\text{V}(w)$ is always 1 when $\text{support\_score}(c, e) = 1$, or, alternatively, by replacing $\text{support\_score}(c_n, e_n) \in \{0, 1\}$ with $\text{support\_score}(c_n,e_n,w_n) \in [0, 1]$.

\subsection{Warrant Generation and Evaluation}
\label{sec:warrrant_gene_eval}

\begin{figure}[t]
    \centering
    \includegraphics[width=1.00\linewidth]{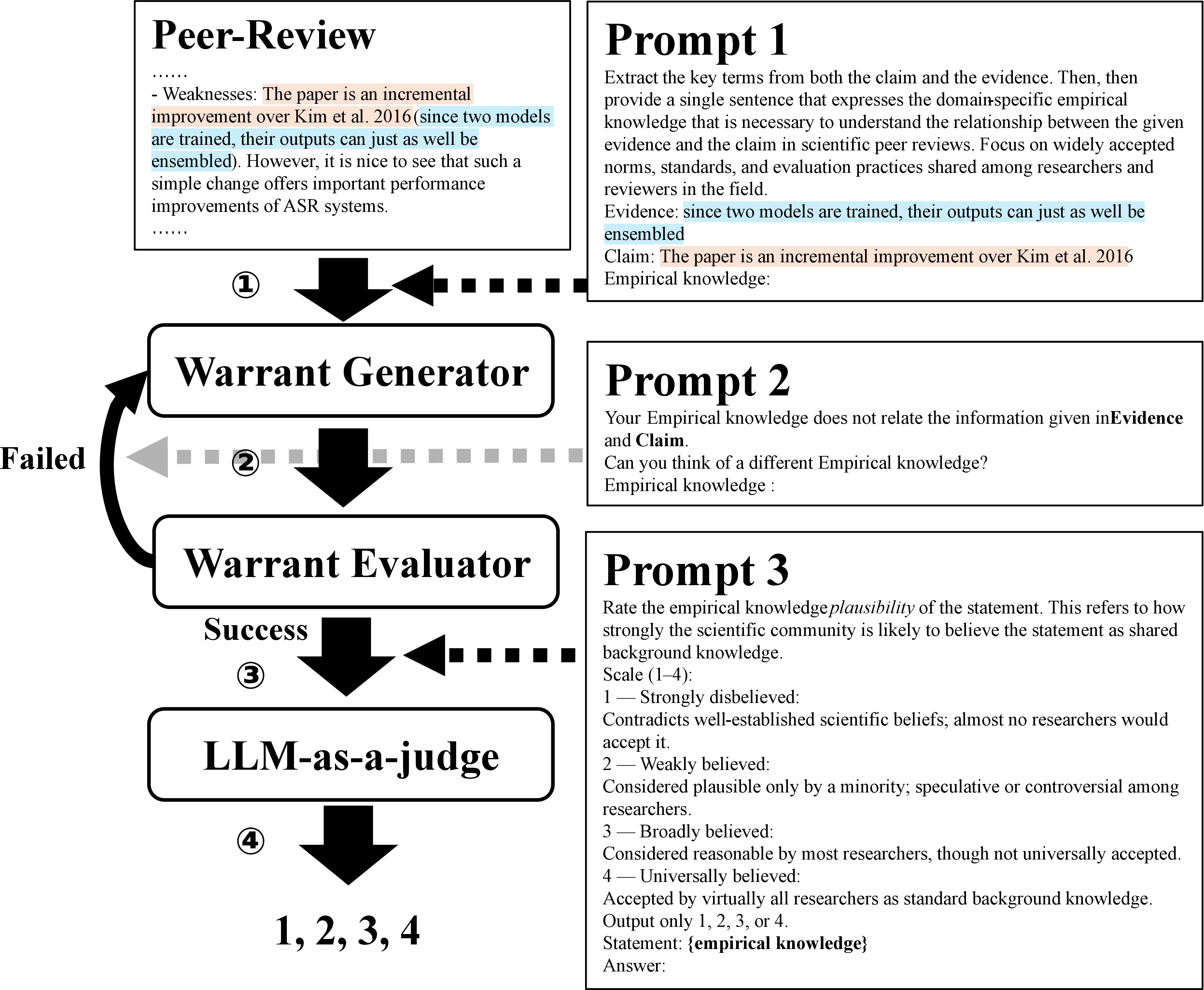} 
    \caption{Warrant-generation and evaluation pipeline using LLMs: first, an LLM (warrant generator) generates a warrant that connects a claim and evidence pair; next, a binary classification model (warrant evaluator) verifies its acceptability; and finally,  another LLM (LLM-as-a-judge) evaluates the warrant on a four-point scale.}
    \label{fig:warrant_generate_evaluate}
\end{figure}

The pipeline for generating warrants from claims and evidence, and evaluating these warrants for the calculation of WarrantScore, is illustrated in Figure \ref{fig:warrant_generate_evaluate}. In our study, a warrant is defined as text that expresses implicit common sense and satisfies the acceptability criteria to connect a claim and its evidence. Since the definition of warrants in Toulmin's model is highly complex, our study focused on generating warrants that are central to evaluation. Specifically, we only focused on the common sense that reviewers have accumulated through their previous reviews or research experience. Details of the implementation for warrant generation and evaluation can be found in Appendix~\ref{sec:warratn_generation_detail} and Appendix~\ref{sec:warrant_evaluation_detail}, respectively.

We use LLM to generate warrants that connect claims and evidence. Then, another LLM classifies the generated warrant as either acceptable or not acceptable using binary classification; if not acceptable, this process is repeated up to three times. If the warrant remains unacceptable after the third classification, we set $w = \emptyset$.

In our study, warrant evaluation is defined as assessing the plausibility of $w$ as $\text{V}(w)$. Using an LLM-as-a-judge, the plausibility of each $w$ is evaluated on a four-point Likert scale. The warrant score $\text{V}(w)$ is derived from a four-point Likert rating $v \in \{1,2,3,4\}$, with the value linearly scaled to the range $[0,1]$ by multiplying by 0.25. In the same way as SubstanScore, if $e$ is $\emptyset$, WarrantScore assigns $\text{V}(w) = 0$. If $e \neq \emptyset$ but $w = \emptyset$, then $V(w) = 1$, following the SubstanScore assumption that any claim supported by evidence is logically connected.

\section{Correlation Analysis with Human Judgments}\label{sec:corr}
In this experiment, we verified that WarrantScore accurately generates warrants that connect claims and evidence, and that it exhibits a stronger correlation with the level of substantiation annotated by humans than SubstanScore. 

\subsection{Evaluation Metrics}
We categorize the evaluation metrics for scientific review comments into our proposed methods and baselines. Our proposed methods, WarrantScore and warrant\_rate, are described in Equation~(\ref{eq:warrant-score}) and Equation~(\ref{eq:warrant-rate}), respectively. For comparison, we also consider the following baseline methods: SubstanScore and supported\_claims, which are defined in Equation~(\ref{eq:substan-score}) and Equation~(\ref{eq:substan-rate}), respectively. We further consider:
\begin{itemize}
    \setlength{\parskip}{0cm} 
    \setlength{\itemsep}{0cm} 
    \item coherence\_rate: The mean semantic similarity, defined as $\frac{\sum_{n=1}^{N} \text{Sem}(c_n, e_n) \times \text{support\_score}(c_n, e_n)}{|C|}$, used to assess the coherence between $c_n$ and $e_n$. \cite{lapata2005automatic}. $\text{Sem}(c, e)$ indicates the calculation of the semantic similarity between $c_n$ and $e_n$ by Sentence-BERT \cite{reimers2019sentence} \footnote{sentence-transformers/all-mpnet-base-v2}.
    \item random$50$\%\_rate: The 50th percentile of the random evaluation that correlate human evaluastion score, calculated by performing the operation $\frac{\sum_{n=1}^{N} r_n \times 0.25 \times \mathrm{support\_score}(c_n, e_n)}{|C|}$ twenty thousand times for all warrants in a dataset, where $r$ is an independently sampled uniform random variable from \{1, 2, 3, 4\} for each $n$.
    \item CoherenceScore: A simple multiple coherence rate and number of words in a review, defined as $ \text{coherence\_rate} \times \text{len(review)}$.
    \item Random$50$\%Score: Analogous to random$50$\%\_rate, but calculated using the formula $\frac{\sum_{n=1}^{N} r_n \times 0.25 \times \mathrm{support\_score}(c_n, e_n)}{|C|} \times \mathrm{len}(\text{review})$, where $\mathrm{len}(\text{review})$.
    \item RM: A regression model based on Llama has been trained using subjective evaluations of the level of substantiation annotated by human. Details of the implementation can be found in Appendix \ref{sec:llm_regression}.
    \item review\_len: The number of words in a review, defined as len(review).
    \item flesch reading ease (fre) \cite{flesch1948new}: a measure of readability given by $\mathrm{FRE}= 206.835- 1.015 \left( \frac{\text{total words}}{\text{total sentences}} \right)- 84.6  \left( \frac{\text{total syllables}}{\text{total words}} \right)$.
    \item type token ration (ttr) \cite{richards1987type}:  measure of vocabulary diversity defined as $\mathrm{TTR} = \frac{V}{\text{total words}}$, where  $V$ is the size of vocabulary within the review.
\end{itemize}
The coherence\_rate and random$50$\%\_rate were used to demonstrate that the correlation between WarrantScore and subjective evaluations of humans  is not attributable to chance. The RM was used as an example of a regression model trained with subjective evaluations of humans  as the target variable, to compare with the scientific review evaluation metrics based on the argumentation structure. To show that changing the warrant evaluation model does not substantially affect the performance of warrant\_rate and WarrantScore, we used two LLM, namely Gemini 2.5 Flash and GPT-5 for warrant evaluation. We also present simple linguistic features, such as review\_len, fre, and ttr, that correlate with aspects of human subjective evaluations of scientific review comments. These features are used to highlight the differences between our proposed scientific review evaluation metric and general linguistic properties.

\subsection{Dataset}
Toward reliable correlation analysis, we used two datasets that were annotated by humans for the level of substantiation: SubstanReview and RottenReview. An overview of each dataset is provided below.

SubstanReview \cite{guo2023automatic} is a dataset in which 550 scientific review comments from the natural language processing scientific review dataset NLPeer \cite{dycke2023nlpeer} are annotated by humans for claims and evidence. Of these, 50 reviews have been assigned the level of substantiation on a three-point Likert scale by three human annotators. In this experiment, only the 50 reviews annotated by humans for their level of substantiation were used to evaluate the proposed evaluation metrics. We aggregated the levels of substantiation assigned by the three annotators by taking the maximum value to reduce variability in scores among human annotators, as evaluating scientific review quality is a challenging task. RM for SubstanReview was not calculated, since the dataset size was insufficient for reliable model training and evaluation, making correlation estimates unstable.

RottenReview \cite{ebrahimi2025rottenreviews} contains 753 scientific review comments randomly sampled from \textit{NeurIPS}, \textit{ICLR}, \textit{F1000}, and \textit{SWJ}. Multiple human annotators evaluated these reviews on a five-point Likert scale across several aspects, including Objectivity, which indicates ``the presence of unbiased, evidence-based commentary.'' Objectivity is a similar aspect to substantiation in SubstanReview. The annotation scores were consolidated by taking the maximum score, because the different annotators evaluated the same review. After deduplication, the number of unique reviews annotated by humans totaled 509 reviews. For this experiment, the correlation between Objectivity and various review evaluation metrics was used to evaluate the metrics. To extract the claim and evidence for all reviews in RottenReview, the state-of-the-art BERT-based model trained on SubstanReview, ModernBERT \cite{warner2025smarter}, was used. Details of the claim and evidence tagging implementation are provided in Appendix \ref{sec:claim and evidence}.

\subsection{Result}
\begin{table}[t]
\centering
\caption{An example of generated warrant in SubstanReview.}
\begin{tabular}{p{1.3cm}|p{5cm}}
\toprule
Claim    & \small{It's great that they also report results on another language,} \\
\hline
Evidence & \small{showing large improvements over existing work on Japanese CCG parsing.} \\
\hline
Warrant  & \small{When a parsing method produces large gains on Japanese CCG and comparable gains on another language, it usually signals language‑agnostic modeling improvements rather than dataset‑specific tricks.}  \\
\bottomrule
\end{tabular}
\label{example_warrant}
\end{table}

\begin{table}[t]
\centering
\small
\caption{Human correlation with peer-review evaluation feature in SubstanReview and RottenReview. The warrants of warrant\_rate and WarrantScore is evaluated by GPT-5.}
\begin{tabular}{lll} 
\toprule
Metric & SubstanReview & RottenReview \\ \midrule
review\_len & \textbf{0.77} & \textbf{0.33} \\
fre & -0.09 & 0.19 \\
ttr & -0.73 & -0.31 \\
\hline
supported\_claims & 0.45 & 0.10 \\
coherence\_rate & 0.67 & \textbf{0.18} \\
warrant\_rate & \textbf{0.69}& \textbf{0.18} \\
\hline
SubstanScore & 0.70 & 0.25 \\
CoherenceScore & 0.79 & 0.25 \\
WarrantScore & \textbf{0.82} & \textbf{0.27} \\
\hline
RM & N/A & \textbf{0.49} \\
\bottomrule
\end{tabular}
\label{general_evaluation_metrics}
\end{table}

\begin{table}[t]
\centering
\small
\caption{Human correlation with warrant\_rate and WarrantScore in SubstanReview and RottenReview. The intersection of the GPT-5 and the rate represents the warrant\_rate evaluated by GPT-5.}
\begin{tabular}{lllll} 
\toprule
Model & \multicolumn{2}{c}{SubstanReview} & \multicolumn{2}{c}{RottenReview} \\
 & rate& score & rate & score \\
\midrule
random50\% & 0.68 & 0.79 & 0.09 & 0.19 \\
\hline
Gemini 2.5 Flash & \textbf{0.69} & 0.80 & 0.16 & 0.26 \\
GPT-5 & \textbf{0.69} & \textbf{0.82} & \textbf{0.18} & \textbf{0.27}\\
\bottomrule
\end{tabular}
\label{warrante_evaluation_metrics}
\end{table}

We evaluated the evaluation metrics for the scientific review by computing Spearman's rank correlation coefficient between subjective evaluations of human and the evaluation metric scores. Table \ref{general_evaluation_metrics} shows the correlation coefficients between them in SubstanReview and RottenReview. In SubstanReview, both warrant\_rate and WarrantScore showed higher correlations with subjective evaluations of humans compared to supported\_claims and SubstanScore, respectively. This suggests that evaluating warrants aligns more closely with human judgments. warrant\_rate and WarrantScore are more strongly correlated with human evaluations than coherence\_rate and CoherenceScore, respectively, suggesting that the explicit generation of warrants leads to better alignment with human judgments. The highest correlation coefficient in SubstanReview was observed for WarrantScore, indicating that when claims and evidence are annotated by humans, WarrantScore most accurately assesses the level of substantiation in scientific reviews.

Similarly, in RottenReview, warrant\_rate and WarrantScore as well as coherence\_rate and CoherenceScore, showed high correlations with subjective evaluations of humans, outperforming supported\_claims and SubstanScore, respectively. Unlike the results with SubstanReview, warrant\_rate and WarrantScore in RottenReview displayed correlation coefficients with subjective evaluations of humans nearly equivalent to correlation coefficients of coherence\_rate and CoherenceScore, which are weaker than the correlation between review\_len and subjective evaluations of humans. Although WarrantScore is more strongly correlated with human subjective evaluations than SubstanScore, the correlation coefficients of both metrics with human evaluations remain relatively low. 

In RottenReview, the evaluation method that showed the highest correlation with subjective evaluation of humans was RM, a regression model. In RottenReview, the assigned scores are not explicitly linked to claim-evidence pairs, and the claim-evidence pairs used in our experiment were automatically extracted. This result highlights a limitation of our approach, which does not directly regress on review scores. However, since our method does not require labeled training data and the evaluation dataset for RM covers only 30\% of the RottenReview, we cannot conclude that our method is inherently inferior.

Table \ref{warrante_evaluation_metrics} shows the correlation coefficients of warrant\_rate, WarrantScore, and random$50$\% for each LLM. In both SubstanReview and RottenReview, when evaluating warrants with LLMs, these metrics exceeded the correlation coefficients obtained from random$50$\%. The increased correlation with human subjective assessments, observed when LLMs evaluate warrants more accurately than random, suggests that proper warrant evaluation is crucial for our proposed method.

An example of claims, evidence, and warrants generated by LLMs in SubstanReview is shown in Table \ref{example_warrant}. Intuitively, in SubstanReview, where claims and evidence are annotated by humans, LLMs appear to generate warrants that connect the given claims and evidence as common sense, functioning as plausible justifications for claims.

\section{Robustness Analysis of Evaluation Metrics}\label{sec:robust}
In this section, we analyzed the advantages of the proposed method, particularly its robustness, by comparing it with a regression model, since the correlation between our proposed method and subjective evaluations of humans is still relatively low. We confirmed that each evaluation metric tends to assign higher scores to good reviews than to bad ones, regardless of the number of words in a review. This property is a desirable characteristic of evaluation metrics for peer-review.

Goldberg et al. \cite{goldberg2025peer} analyzed the reliability of scientific review comment evaluations by presenting experts with review comments submitted to international conferences, along with artificially elongated versions of those reviews that retain the original content but include redundant text. The results indicated that experts tend to evaluate elongated reviews more favorably than the original reviews. This highlights the need for evaluation metrics that accurately evaluate the quality of scientific reviews while being unaffected by biases introduced by number of words in a review, unlike subjective evaluations of human.

The extent to which review evaluation metrics are influenced by number of words in a review was assessed using the following two datasets. 

One dataset confirms that scientific review evaluation metrics tend to assign lower scores to deficient reviews compared to sufficient reviews. This dataset is referred to as ReviewGuardReview. ReviewGuard \cite{zhang2025reviewguard} is a framework that uses an LLM-as-a-judge to evaluate the sufficiency of reviews, labeling reviews as sufficient or deficient. ReviewGuard defines a review as sufficient if it meets all three of the following criteria: a) commitment (the extent of substantive engagement with the manuscript), b) constructive intention (a fair-minded desire to improve the work), c) domain knowledgeability (sufficient expertise to make an informed judgment). Missing any of them makes it deficient.

In their study, following the ReviewGuard framework, labels were assigned to all reviews of \textit{ICLR 2019} using GPT-OSS. Subsequently, 500 reviews labeled as sufficient and 500 reviews labeled as deficient were randomly sampled to construct the dataset. Note that only the review texts were used as input; the paper's title and abstract, which are included in the original framework, were omitted, as the review evaluation metrics do not reference these details. In our study, we assume that sufficient reviews are well substantiated, while deficient reviews are poorly substantiated.

\begin{figure}[t]
    \centering
    \includegraphics[width=1\linewidth]{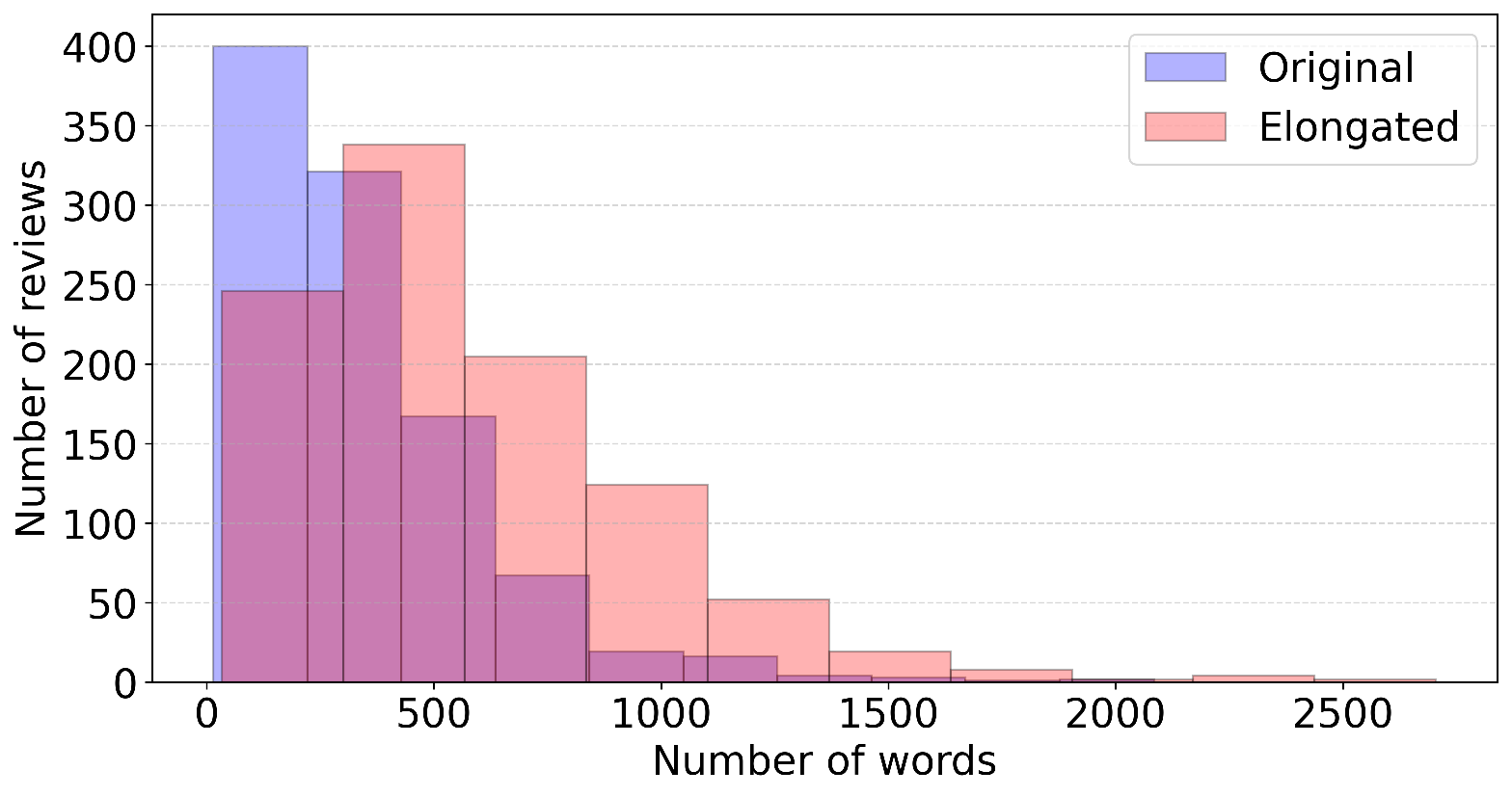} 
    \caption{Number of words in the original peer-review and the elongated peer-review.}
    \label{fig:reviewlen}
\end{figure}
Another dataset is used to verify whether scientific review evaluation metrics assign similar scores to original and elongated reviews. This dataset is referred to as ElongatedReview. ElongatedReview is inspired by the experiment conducted by Goldberg et al. \cite{goldberg2025peer}, which showed that experts tend to rate elongated reviews more highly. ElongatedReview is created by taking 1,000 reviews from ReviewGuardReview and generating elided summaries of each review using GPT-OSS, then appending these summaries to the original reviews. Consequently, for each of the 1,000 reviews, a nearly identical but more elongated version was produced. Figure \ref{fig:reviewlen} shows the word counts for both the original and elongated reviews. The average word count of the original reviews is 337.47 words, with the minimum being 15 words and the maximum 2,084 words. For the elongated reviews, the average word count is 572.26 words, with the minimum at 34 words and the maximum at 2,705 words.

\begin{table}[t]
\centering
\scriptsize
\caption{The mean evaluation score of each feature (e.g., warrant\_rate) for peer-review comments labeled as sufficient or deficient reviews in ReviewGuardReview. The warrants of warrant\_rate and WarrantScore is evaluated by GPT-5.}
\begin{tabular}{lllll} 
\toprule
Metric & sufficient & deficient &d($\uparrow$) &p($\downarrow$) \\ \midrule
review\_len & \textbf{366.78} & 308.16 & \textbf{0.22} & \textbf{0.00} \\
fre & \textbf{41.72} & 41.25 &0.04& 0.26 \\
ttr & 0.55 & \textbf{0.59} &-0.37& 1.00 \\
\hline
supported\_claims & 0.44 & \textbf{0.46} &-0.05& 0.81 \\
coherence\_rate & \textbf{0.11} & 0.09 &\textbf{0.21}& \textbf{0.00} \\
warrant\_rate & \textbf{0.28} & 0.23 & 0.18 & \textbf{0.00}\\
\hline
SubstanScore & \textbf{175.09} & 150.07 &0.14& \textbf{0.01} \\
CoherenceScore & \textbf{48.11} & 38.38 &\textbf{0.17} & \textbf{0.00} \\
WarrantScore & \textbf{115.38} & 95.25 & 0.16 & \textbf{0.01}  \\
\hline
RM & 1.59 & \textbf{1.77} &-0.15& 0.99 \\
\bottomrule
\end{tabular}
\label{normal_evaluation_metrics_reviewguard}
\end{table}

\begin{table}[t]
\centering
\scriptsize
\caption{The mean evaluation score of each feature (e.g., warrant\_rate) for peer-review comments labeled as original and elongated peer reviews in ElongatedReview. The warrants of warrant\_rate and WarrantScore is evaluated by GPT-5.}
\begin{tabular}{lllll} 
\toprule
Metric & original & elongated & d ($\downarrow$) & p($\uparrow$) \\ \midrule
review\_len & 337.47 & \textbf{572.26} & 1.99  & 0.00\\
fre & \textbf{41.49} & 31.96 & \textbf{-1.84} & 0.00\\
ttr & \textbf{0.57} & 0.47 &-2.17 & 0.00 \\
\hline
supported\_claims & 0.45 & \textbf{0.47} & \textbf{0.07} & \textbf{0.03} \\
coherence\_rate & 0.10 & \textbf{0.11} & 0.11 & 0.00 \\
warrant\_rate & 0.26 & \textbf{0.27} & \textbf{0.07} & \textbf{0.02}\\
\hline
SubstanScore & 162.58 & \textbf{284.66} &0.89 & 0.00\\
CoherenceScore & 42.34 & \textbf{72.12}& 0.69 & 0.00\\
WarrantScore & 105.32 & \textbf{174.91} & \textbf{0.68} & 0.00\\
\hline
RM & \textbf{1.68} & 1.05 &-0.77&0.00\\
\bottomrule
\end{tabular}
\label{warrant_evaluation_metrics_normal}
\end{table}

\subsection{Evaluation Metric}
In the evaluation using ReviewGuardReview, a one-sided independent t-test was conducted with the null hypothesis that the sufficient reviews are not rated higher than the deficient reviews. If the null hypothesis is rejected, it indicates that the sufficient reviews are rated higher than the deficient ones. In ElongatedReview, a paired t-test was used with the null hypothesis that the evaluation scores of the original reviews and the elongated reviews are the same. If the null hypothesis is not rejected, it suggests that the original and elongated reviews are evaluated as equivalent. Both tests in the dataset are significant at the 1\% level. The effect sizes for each comparison were calculated using Cohen's $d$ \cite{cohen2013statistical}.
For ReviewGuardReview, $d$ was computed as the difference between sufficient and deficient reviews (sufficient $-$ deficient).
For ElongatedReview, $d$ was computed as the difference between the evaluation scores of the original reviews and the elongated reviews (original $-$ elongated).

For the calculation of WarrantScore and SubstanScore, all review comments in datasets were automatically tagged with claims and evidence using ModernBERT. In this experiment, desirable evaluation metrics were those that assign higher scores to sufficient reviews compared to deficient ones, and these scores should be invariant with respect to the number of words in a review.

\subsection{Result}
The evaluation results of the scientific review metrics in ReviewGuard are shown in Table \ref{normal_evaluation_metrics_reviewguard}. The null hypothesis that sufficient reviews are rated lower than deficient reviews was rejected for the following metrics: review\_len, coherence\_rate, warrant\_rate, SubstanScore, CoherenceScore, and WarrantScore. These indicators tend to assign higher scores to sufficient reviews compared to deficient ones. In ReviewGuardReview, RM, the regression method based on the highest correlation coefficient with subjective evaluations of humans does not show a tendency to rate sufficient reviews higher than deficient reviews. Conversely, warrant\_rate, which has limited correlation with subjective evaluations of humans, can still rate sufficient reviews higher than deficient ones. This suggests that the overall rankings of scientific review evaluation metrics cannot be determined solely based on their correlation with subjective evaluations of humans.

The evaluation results for scientific review evaluation metrics based on ElongatedReview are shown in Table \ref{warrant_evaluation_metrics_normal}. The null hypothesis that original reviews and elongated reviews are evaluated equally was not rejected for the following metrics: supported\_claims and warrante\_rate. These indicators tend to assign similar scores to original reviews and elongated ones. In other words, in terms of robustness, supported\_claims and warrant\_rate have an advantage over other features.

According to the experimental results, only the warrant\_rate, a core component of WarrantScore, provides a robust evaluation of peer-review comments, as the null hypothesis that sufficient reviews are rated lower than deficient reviews was rejected, whereas the null hypothesis that original and elongated reviews are evaluated equally was not rejected.

\section{Conclusion}
In this study, we proposed a new framework to support the evaluation of scientific reviews, with a particular focus on claim–evidence pairs and their logical inferences. Specifically, we introduced WarrantScore, a metric for evaluating peer-review comments that considers not only supporting evidence but also the plausibility of warrants connecting claims and evidence. Experiments on human-annotated datasets show that WarrantScore correlates more strongly with human judgments than conventional metrics, indicating that modeling warrants enhances both alignment with human evaluations and the interpretability of review assessments.

\clearpage
\section*{Limitations}
One limitation of our study is that we fixed the warrant, the rationale connecting claims and evidence, to a single form and limited the types of warrants to common sense likely acquired by reviewers through past reviews and research experience. Warrants are not necessarily unique, and their types are diverse. Consequently, the evaluation value of an arbitrary warrant is likely non-unique. In our proposed approach, we evaluated candidate warrants for each claim and evidence pair as a four-point Likert scale, using an LLM-as-a-judge. Although the generated warrants are understandable to humans, interpreting the evaluation scores assigned to these warrants remains challenging. Our primary demonstration is that, by leveraging the concept of warrants, generating and assessing implicit text that connect claims and evidence, we can improve the correlation between scientific review evaluation metrics and subjective evaluations of human, as well as the interpretability of the evaluation scores themselves.

Another limitation of our study concerns the small dataset size and the differences in domains. We used two publicly available human-annotated datasets, as well as two datasets generated with large language models. Our proposed method shows consistent performance improvements over conventional methods on all datasets, but the actual effectiveness remains unclear. Clarifying the practical performance of our approach requires further development of a large, human-annotated datasets of scientific reviews.

\section*{Acknowledgments}
This work was supported by JST
Moonshot R\&D Program, Grant Number JPMJMS2236.
\section*{Ethics Statements}
AI Assistants in our study including ChatGPT and Google Gemini, were used in accordance with the ACL Policy on AI Writing Assistance. We primarily used them to assist with
coding and writing, but all code and text outputs
were manually reviewed. The authors take full
responsibility for all of them.

\bibliography{custom}

\appendix

\section*{Appendix}

\section{Detail of Warrant Generation and Evaluation}
\label{sec:warratn_generation_detail}
For the generation of warrants, we utilized GPT-OSS \footnote{openai/gpt-oss-120b}, a large open-source LLM, with the temperature set to 0. When regenerating warrants, the previously generated warrants are provided as context. During warrant evaluation, the maximum number of output tokens generated by the LLM was set to 2, and the parameter do\_sample was set to false.
\section{Warrant Acceptability Assessment}
\label{sec:warrant_evaluation_detail}

To estimate whether the warrants generated by the LLM are acceptable, we employed a binary classification model based on Llama \footnote{meta-llama/Meta-Llama-3-8B}. We trained this model using a dataset containing 450 warrants with human subjective acceptability annotations, which was publicly released by Gupta et al. \cite{gupta2024harnessing}. During training, the inputs to Llama consisted of claims, evidence, warrants, and the target variable–human annotations of warrant acceptability. The model was fine-tuned with LoRA \cite{hu2022lora} for binary classification. The data was split into 70\% training and 30\% evaluation sets. During training, the hyperparameters were: 5 epochs, batch size of 4, learning rate of 5e-5, maximum input sequence length of 2048 tokens, LoRA rank of 16, scaling coefficient of 32, and dropout rate of 0.05. The modules where LoRA was applied were q\_proj and v\_proj. The trained model achieved an accuracy of 91.80\%, a recall of 83.58\%, and an F1 score of 87.5\% on the evaluation dataset. Therefore, the trained model can at least reliably judge the acceptability of warrants in a general domain.

\section{Training Setting for Regression Model}
\label{sec:llm_regression}

For implementing the evaluation metric, we trained a regression model using Llama, with the input being review texts and human subjective evaluations of Objectivity from RottenReview. The training data comprised 70\% of the dataset, and the evaluation data 30\%. During training, the hyperparameters were set as follows: 5 epochs, a learning rate of 5e-5, a batch size of 8, and a maximum input length of 8,096 tokens. Regarding LoRA, the rank was set to 16, the scaling coefficient to 32, and the dropout rate to 0.05, applied to the q\_proj and v\_proj modules.

\section{claim and evidence extract}
\label{sec:claim and evidence}
To implement SubstanScore, we replicated the claim tagging and claim and evidence linking experiments conducted by Guo et al. \cite{guo2023automatic} using LLMs, with modifications to the training conditions for this specific experiment.

Claim tagging involves a token classification task \cite{segal2020simple} that extracts multiple spans containing claims from a given review text. Linking evidence involves a Question Answering (QA) extraction task \cite{rajpurkar2016squad} that extracts corresponding evidence spans from claim-review pairs. In our study, we used the state-of-the-art BERT-based model, ModernBERT \cite{warner2025smarter}\footnote{answerdotai/ModernBERT-large}, which supports a maximum input token length of 8,096 tokens, to replicate both claim tagging and evidence linking experiments.

The claim and evidence span in the scientific review texts for this experiment were cropped to remove punctuation and whitespace, including spaces, tabs, and newlines, from both ends of the spans. These spans were derived from claim and evidence annotations manually created in SubstanReview. The training dataset comprised 440 reviews, and the evaluation dataset comprised 110 reviews. We combined all training and evaluation data and conducted 5-fold cross-validation to evaluate the model's performance.

\begin{table}[t]
\centering
\scriptsize
\caption{Comparison of claim and evidence pair extraction results between state-of-the-art models in related work and our model.}
\begin{tabular}{llllll} 
\toprule
 & \multicolumn{3}{c}{Claim tagging} & \multicolumn{2}{c}{Evidence linking} \\ 
 & Precision & Recall & F1 & EM & F1 \\ \midrule
Baseline & 52.00 & 59.77 & 55.61 & 64.31 & 82.07 \\
Our & \textbf{61.36} & 53.13 & \textbf{56.91} & \textbf{66.68} & 71.98 \\
\bottomrule
\end{tabular}
\label{table:claim and evidence}
\end{table}

\subsection{Claim Tagging}
The hyperparameters used for training the claim tagging model are as follows: batch size of 8, 20 epochs, learning rate of 4e-5, weight decay of 0.10, maximum gradient norm of 1.0, hidden layer dropout rate of 0.1, and attention mechanism dropout rate of 0.1. In prior research, positive and negative claims were distinguished by applying BIO (Beginning, Inside, Outside) encoding, training the token classification model to identify five classes: B-claim\_positive, I-claim\_positive, B-claim\_negative, I-claim\_negative, and O. However, since SubstanScore does not differentiate between positive and negative claims, our study simplifies the task by combining both into a single ``claim'' category for tagging. Consequently, the classes for token classification are B-Claim, I-Claim, and O. The evaluation of claim tagging and the span annotations in subsequent tasks define a span as the sequence starting from a B-Claim token, continuing with consecutive I-Claim tokens, and ending when O tokens appear.

\subsection{Evidence Linking}
The training for the evidence linking task involves inputting the claim and review sentences $R = \{ r_1, r_2, \ldots, r_{|R|} \}$ along with the claims $C = \{ c_1, c_2, \ldots, c_{|C|}\}$ as a combined sequence structured as $\{ [CLS] c_1 c_2 \cdots c_{|C|} [SEP] r_1 r_2 \cdots r_{|R|} \}$. This input format is used to extract the span of evidence corresponding to each claim. The hyperparameters during training are batch size of 4, 16 epochs, learning rate of 1e-5, mixed precision training enabled via FP16, with a hidden layer dropout rate of 0.2 and an attention mechanism dropout rate of 0.2.

claim and evidence extraction is evaluated using the same metrics as in prior research, with the results shown in Table \ref{table:claim and evidence}. These results demonstrate that the pre-trained ModernBERT model achieves performance comparable to the state-of-the-art models reported in previous studies. Therefore, for claim and evidence extraction in our experiments, we utilize this trained ModernBERT. The maximum input token length is set to 8,096 tokens, and the model is trained on the 440 reviews from the SubstanReview training dataset.

\end{document}